\documentclass[10pt,twocolumn,letterpaper]{article}

\usepackage{iccv}
\usepackage{times}
\usepackage{epsfig}
\usepackage{graphicx}
\usepackage{amsmath}
\usepackage{amssymb}

\usepackage{color,xcolor}
\usepackage{booktabs}
\usepackage{multirow}
\usepackage[accsupp]{axessibility}

\definecolor{citecolor}{rgb}{0.0, 0.56, 0.0}
\definecolor{algcolor}{rgb}{0.0, 0.3, 0.0}

\usepackage[pagebackref=False,
breaklinks=true,
colorlinks=true,
citecolor=citecolor,
linkcolor=red,
bookmarks=false]{hyperref}

\newcommand*{\Video}{\mathcal{V}}
\newcommand*{\Image}{\mathcal{I}}

\iccvfinalcopy 

\def\iccvPaperID{3110} 

\ificcvfinal\fi

\begin{document}

\title{Cross-view Semantic Alignment for Livestreaming Product Recognition}

\author{
  Wenjie Yang\thanks{Equal contribution}~, Yiyi Chen\footnotemark[1]~, 
  Yan Li, Yanhua Cheng,
  Xudong Liu, Quan Chen\thanks{Corresponding author}~, 
  Han Li\\
  Kuaishou Technology\\
  \footnotesize\texttt{wenjie.yang@nlpr.ia.ac.cn},
  \footnotesize\texttt{\{chenyiyi,liyan26,chengyanhua,liuxudong,chenquan06,lihan08\}@kuaishou.com}
}



\maketitle
\ificcvfinal\thispagestyle{empty}\fi

\begin{abstract}
    Live commerce is the act of selling products online through live streaming.
    The customer's diverse demands for online products introduce more challenges to Livestreaming Product Recognition. 
    Previous works have primarily focused on fashion clothing data or utilize 
    single-modal input, which does not reflect the real-world scenario where
    multimodal data from various categories are present.
    In this paper, we present LPR4M,
    a large-scale multimodal dataset that covers 34 categories, comprises
    3 modalities (image, video, and text), and is 50$\times$ larger than the
    largest publicly available dataset.
    LPR4M contains diverse videos and noise modality pairs while
    exhibiting a long-tailed distribution, resembling real-world problems.
    Moreover,~a
    c\textbf{R}oss-v\textbf{I}ew semanti\textbf{C} alignm\textbf{E}nt (RICE) model
    is proposed to learn discriminative instance features from the image and video views of the products. This is achieved through instance-level
    contrastive learning and cross-view patch-level feature propagation.
    A novel Patch Feature Reconstruction loss is proposed to penalize the semantic
    misalignment between cross-view patches.
    Extensive experiments demonstrate the
    effectiveness of RICE and provide insights into the importance of dataset diversity and expressivity.
    The dataset and code are available at \url{https://github.com/adxcreative/RICE}.
   
\end{abstract}

\section{Introduction}
\begin{figure}[htp]
    \begin{center}
    	\includegraphics[width=0.90\linewidth]{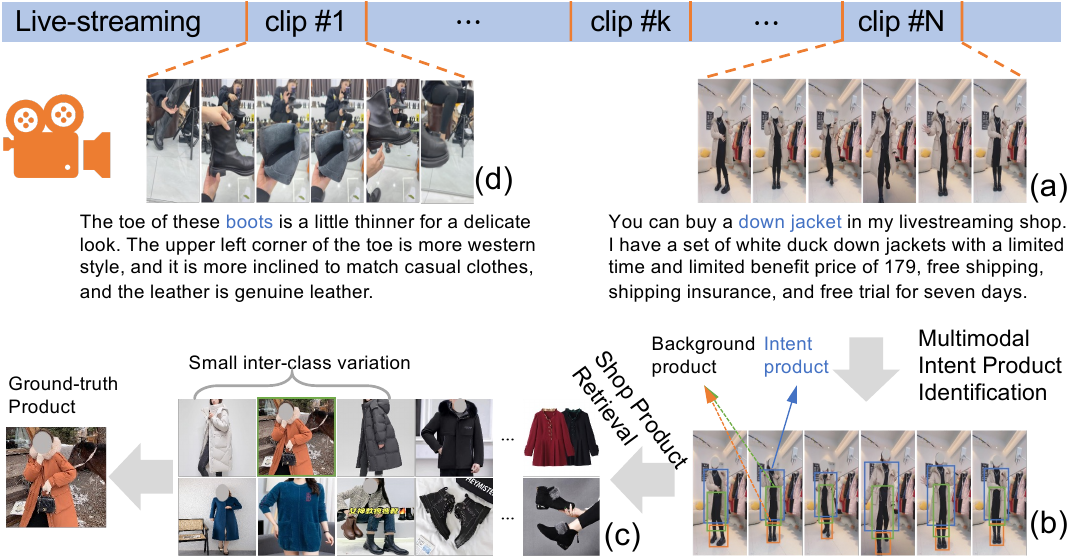}
    \end{center}
    \vspace{-0.9em}
    \caption{The pipeline of LPR. A livestreaming consists of many clips introducing different products. 
    We show two clip examples with ASR text in (a) and (d).
    In (b), the intended product refers to the product the salesperson is introducing, and the other products on the screen are indicated as the distracted background products. 
    (c) presents a shop with hundreds of images, some with subtle visual differences called small inter-class variations. The LPR aims to identify the clip's intended product using the ASR text prompt, then retrieve the ground-truth product from the shop images.
    }
    \vspace{-1.5em}
    \label{fig:intro}
\end{figure}



Livestreaming Product Recognition (LPR)~\cite{cheng2017video2shop, godi2022movingfashion, hadi2015buy} is one of the significant machine learning application in the e-commerce industry. Its goal is to recognize products a salesperson presents in a live commerce clip through content-based video-to-image retrieval. The real-time and accurate recognition of livestreaming products can facilitate the online product recommendation, and thereby improve the purchasing efficiency of consumers.

\begin{figure*}[htp]
    \begin{center}
    \includegraphics[width=0.85\linewidth]{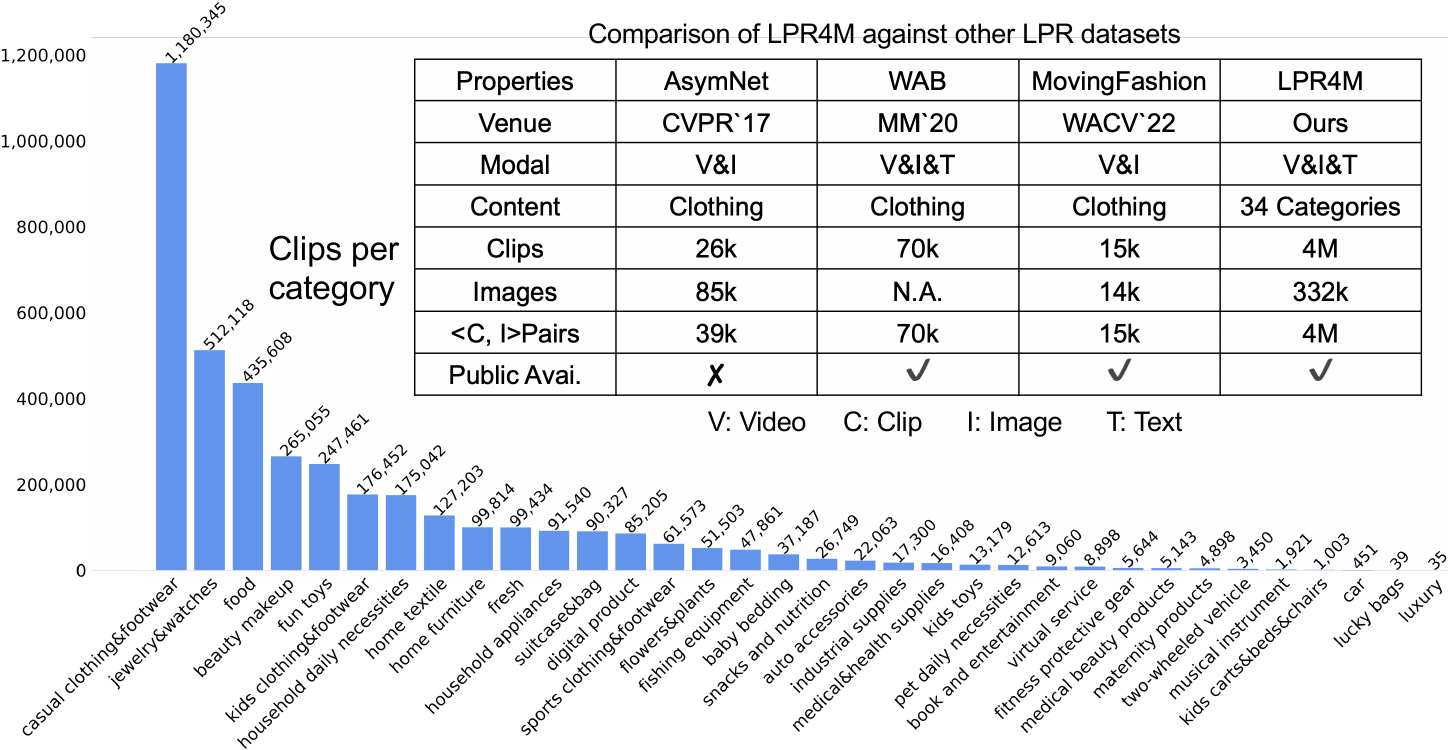}
    \end{center}
    \vspace{-1.5em}
    \caption{BAR CHART: Number of clips per category for LPR4M, with a long-tailed distribution resembling most real-world problems.
    TABLE: Comparison of LPR4M against other LPR datasets in terms of 
    modal, content, and scale.
    LPR4M offers significantly broader coverage of live commerce product categories and several orders of magnitude larger data scales.
    }
    \label{fig:cate2vid}
    \vspace{-1.5em}
\end{figure*}

The task of LPR involves two fundamental processes: multimodal-based intended product identification and shop product retrieval. This task poses significant challenges in real-world scenarios, including 
(1) the need to distinguish \textit{intended products} from the cluttered background products in a livestreaming frame, exemplified in Fig.~\ref{fig:intro} (b), 
(2) the requirement for models to capture sufficient \textit{fine-grained} features to match the ground-truth (GT) image accurately in the shop, where there are many images with subtle visual nuances, 
(3) the \textit{heterogeneous} video-to-image and \textit{cross-domain} livestream-to-shop problem, and 
(4) the \textit{appearance changes} of products in the livestreaming domain due to articulated deformations, occlusions, diverse background clutters, and significant illumination variations, making it a highly intricate task to match the clip to the GT image in the shop.
Various datasets have emerged in the computer vision community to study this task, including AsymNet~\cite{cheng2017video2shop}, WAB\footnote{https://tianchi.aliyun.com/competition/entrance/231772/information}, and MovingFashion~\cite{godi2022movingfashion}. However, AsymNet and MovingFashion lack crucial text modal, which provides essential auxiliary information for identifying intended products. Furthermore, the data scale of WAB is relatively small, with only 70K pairs, and only provides fashion clothing data, diverging from the real-world scenario.
 

In order to narrow the gap between existing datasets and the real-world scenario and advance research in this challenging task, we present LPR4M, a large-scale multimodal live commerce dataset that includes extensive categories, diverse data modalities of clip, image, and text, as well as heterogeneous and cross-domain correspondences of $\langle clip, image\rangle$ pairs. This dataset offers several significant advantages. (1) \textit{Large-Scale:} LPR4M contains over 4M pairs, significantly exceeding its precedents. 
(2) \textit{Expressivity:} LPR4M draws data pairs from 34 commonly used live commerce categories rather than relying solely on clothing data. Additionally, LPR4M offers auxiliary clip ASR text and image title modalities, which are critical for intended product identification and product feature representation. 
(3) \textit{Diversity:} LPR4M promotes clip diversity while preserving the real-world data distribution, with a focus on three components: product scale, visible duration, and the number of products in the clip, as depicted in Fig.~\ref{fig:example}. To the best of our knowledge, LPR4M is currently the largest dataset created explicitly for real-world multimodal LPR scenarios.



Our work based on LPR4M tackles a realistic problem: \textit{how to achieve fine-grained LPR using large-scale multimodal pairwise data?} Given image and clip views, we first utilize Instance-level Contrastive Learning (ICL) to align global features. However, since instance features of these two views are extracted independently from the visual encoder, it can be challenging to differentiate between products with subtle visual differences without cross-view interactions. Consequently, we propose a patch-level semantic alignment approach to enable cross-view patch information propagation. We suggest measuring similarity via a cross-attention based Pairwise Matching Decoder (PMD), which treats image patches as \textit{Query} and video patches as both \textit{Key} and \textit{Value}. In addition, we propose a novel Patch Feature Reconstruction (PFR) loss to provide patch-level supervision for pairwise matching, expecting to reconstruct each feature of an image patch from its paired video patches.

The main contributions of this paper can be summarized as follows.
(1) A large-scale live commerce dataset is collected, offering a significantly broader coverage of categories and diverse modalities such as video, image, and text.
This dataset is the most extensive one known to date, tailored explicitly for real-world multimodal LPR scenarios. 
(2) The RICE model is introduced to integrate instance-level contrastive representation learning and patch-level pairwise matching into a framework.
(3) A novel Patch Feature Reconstruction loss is proposed to penalize the semantic
misalignment between patches of video and image.
(4) The benchmark dataset and evaluation protocols are carefully defined 
for LPR. Extensive experiments demonstrate the effectiveness of LPR4M and RICE.

\section{Related Works}

\textbf{LPR datasets.}
As shown in the table of Fig.~\ref{fig:cate2vid}, we compare LPR4M with others in terms of modality, content, and scale. 
In particular, AsymNet~\cite{cheng2017video2shop}, WAB, and MovingFashion~\cite{godi2022movingfashion} only provide fashion clothing data, and the text modality is absent in AsymNet and MovingFashion. Therefore, 
We collect LPR4M, 
which covers 34 widely used categories and provides visual and text modalities.
LPR4M is 50$\times$ larger than WAB.

\textbf{Video Object Detection (VOD).} The main focus of recent VOD
methods~\cite{zhu2017deep,zhu2017fgfa, deng2019rdn, chen2020mega, sun2021mamba,zhou2022transvod} is exploiting temporal information to tackle the video variations, \eg, occlusion, motion blur and out of focus.
The temporal relationship mining insights in VOD inspire this paper's design of the intended product detection module.
However, unlike VOD, where most videos only contain a single object, the videos in LPR contain many cluttered background products. 
This situation suggests that in the absence of prompt text information, it is more challenging to identify the intended product by only relying on the visual inputs.
Therefore, we explore the fusion of text and visual modalities and verify its effectiveness in this paper.

\textbf{Fine-Grained Vision Recognition (FGVR).}
Similar to FGVR~\cite{fu2017look,zheng2019looking,ding2019selective,li2018discriminative,ge2019weakly,zhu2022dual,yang2019towards,yang2022bottom,bai2023cross}, the LPR aims to learn discriminative instance feature to 
distinguish the subclasses with large intra-class and small inter-class variations.
However, unlike traditional FGVR, each live commerce category contains an enormous amount of subclasses in LPR. Moreover, the number of subclasses will increase or decrease dynamically as a large number of products are newly added or taken off the shop every day.
It makes it more challenging to handle the out-of-distribution subclasses.

\textbf{Video-to-Shop Retrieval.}
Although fashion retrieval has made great progress~\cite{huang2015cross, liu2016deepfashion,ge2019deepfashion2,kuang2019fashion},
there are few studies focus on retrieving products that are presented in e-commerce
videos, referred to as \textit{video-to-shop}.
AsymNet is a \textit{one-stage} method without detection. It employs LSTM to exploit temporal continuity in the video, then perform pair-wise matching by feeding the image and video feature into a similarity network.
DPRNet~\cite{zhao2021dress} and SEAM Match-RCNN~\cite{godi2022movingfashion} adopt a \textit{two-stage} pipelines.
DPRNet first detects the products in the video and then performs image-to-image retrieval.
SEAM Match-RCNN performs self-attention among the detected product boxes in a video to produce a video feature and uses the inner product between the video and image feature as a similarity.
In this paper, we propose RICE integrates the \textit{one-stage} and \textit{two-stage} methods into a framework and study their advantages. 

\section{Dataset and Benchmark}
In this section, we present the construction, characteristics, and benchmark of 
LPR4M. 

\begin{figure*}[htp]
    \begin{center}
    	\includegraphics[width=0.85\linewidth]{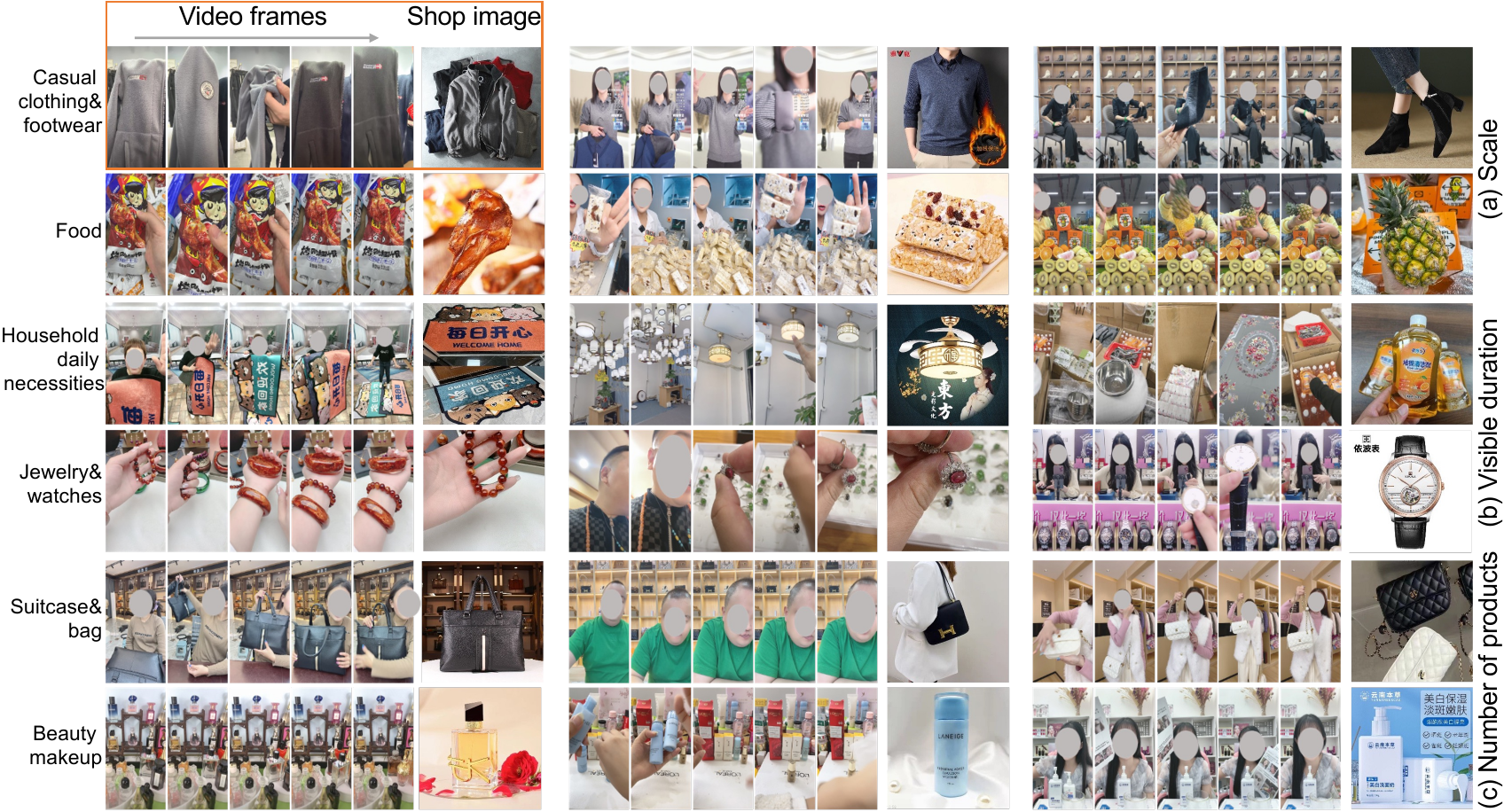}
    \end{center}
    \vspace{-1.0em}
    \caption{The $<clip, image>$ pairs of LPR4M. 
    As shown on the left of the orange box, we extract five evenly spaced frames from the clip, with the shop image on the right.
    We choose two categories to illustrate one of the clip product variations of \textit{scale}, \textit{visible duration}, and \textit{number of products}. 
    Each row shows 3 data pairs for different degrees of difficulty of the corresponding variation, including (a) \textit{large}, \textit{medium} and \textit{small} product scale, (b) \textit{long}, \textit{medium} and \textit{short} visible duration, (c) \textit{abundant}, \textit{medium} and \textit{few} products in the clip.
    }
    \vspace{-1.0em}
    \label{fig:example}
 \end{figure*}

\begin{figure*}[t]
    \begin{center}
    	\includegraphics[width=0.85\linewidth]{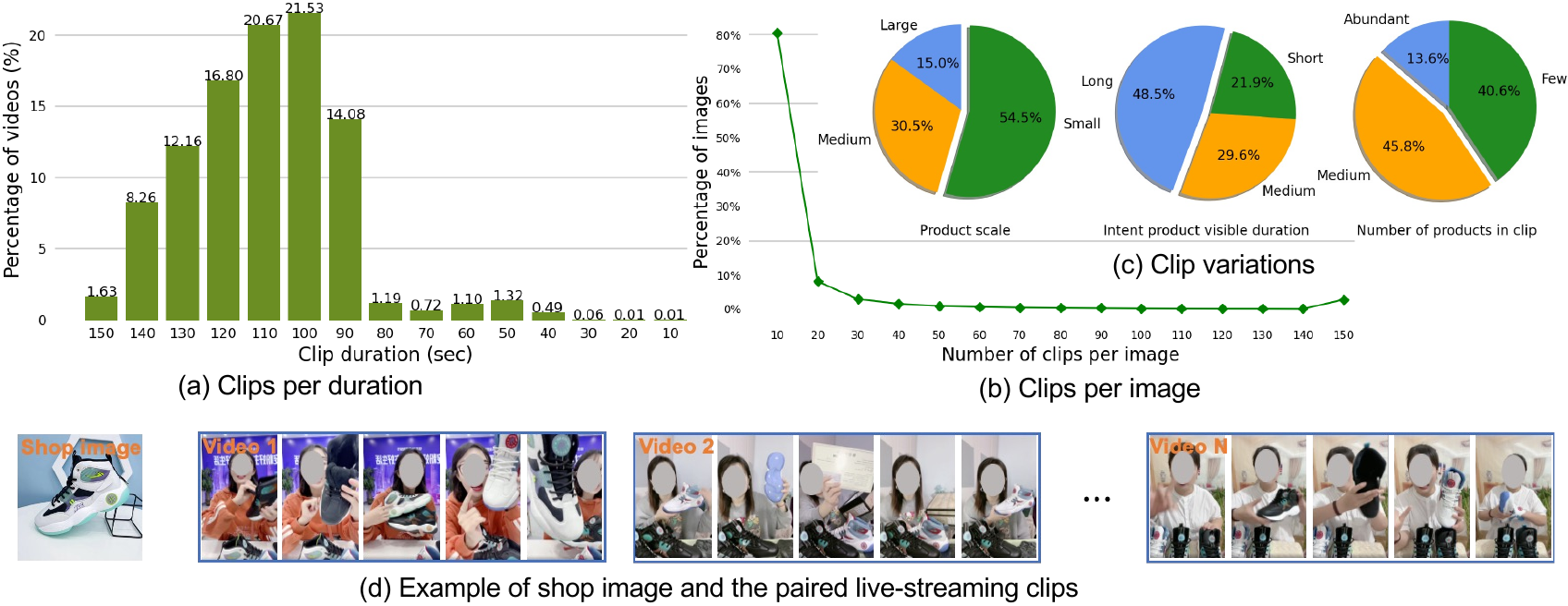}
    \end{center}
    \vspace{-1.0em}
    \caption{LPR4M statistics of clips. (a) The distribution of clip duration. (b) The number of clips per image. Approximately 80\% of the images have ten paired clips. (c) The statistics of three clip variations.
    (d) The shop image, the paired clips, and the products in the clips suffer from different variations, \eg, scale, viewpoint, and occlusion.
    }
    \vspace{-1.6em}
    \label{fig:analysis}
 \end{figure*}

\textbf{Overview.}
Compared with other existing LPR datasets, LPR4M has several appealing properties, which are summarized in the following.
(1) \textit{Large-Scale.} 
As illustrated in the table of Fig.~\ref{fig:cate2vid}, LPR4M is the largest LPR dataset to date. 
It contains 4M exactly matched $\langle clip, image\rangle$ pairs of 4M live clips, and 332k shop images. Each image has 14.5 clips with different
product variations, \eg, viewpoint, scale, and occlusion. The example of 
image-to-clips and the number of clips per image are shown in Fig.~\ref{fig:analysis} (d) and (b), respectively. Specifically, most of the
images (80\%) have ten matched clips and the number of clips per image range
from 10 to 150. 
(2) \textit{Expressivity.}
The expressivity of LPR4M is mainly reflected in two aspects. Firstly, unlike other LPR datasets that only contain fashion clothing data, our data is more affluent, coming from 34 categories covering most of the daily necessities. This makes it closer to the real scenario.
Secondly, the data of LPR4M is multimodal. We provide live clip ASR texts and shop image titles as auxiliary information to facilitate the intended product identification and form a full-scale characteristic of each product.
(3) \textit{Diversity.}
Firstly, we collect clips according to the clip duration distribution of real livestreaming scenarios and obtain the clips with various durations, as shown in Fig.~\ref{fig:analysis} (a).
Secondly, the clips are further sampled by controlling the
variation in terms of three properties, \ie, product scale, intended product
visible duration, and the number of products in the clip. It makes LPR4M a challenging benchmark. 
As illustrated in Fig.~\ref{fig:example}, we pick two categories to represent
a variation. For each category row, the clip shows three different levels of
difficulty progressively.

\subsection{Data Collection and Cleaning} 

The basic unit of the dataset is a $\langle clip, image\rangle$ pair.
All the clips are cut from hours of sequential livestreaming data 
crawled from Kuaishou\footnote {https://live.kuaishou.com}.
A livesteaming has a unique online shop, which lists all the products to be introduced in this livestreaming. 
Firstly, we removed near- and exact-duplicate images in the shops by comparing 
the global average pooled \textit{layer4} features after feeding them into ResNet~\cite{he2016deep}. 
Secondly, the human annotators cleaned the clips that contain the target product 
with short visible duration, small scales, and severe background clutters.
Finally, given a clip, the matched product image is picked from the shop via
the human annotator.
In total, 4,033,696 clips and 398,796 images are kept to construct the training and test set of LPR4M.

\textbf{Variations.} 
The proportion of the number of clips in each variation is depicted 
in Fig.~\ref{fig:analysis} (c).
(1) \textit{Scale.} According to the proportion (\textit{p}) of the product box area to 
the entire frame area, the clips are classified into three subsets. 
The area is measured as the number of pixels in the product box.
In LPR4M, there are more small products than large products. Specifically, approximately 54.5\% of products are small (\textit{p}$\leq$0.2), 30.5\% are medium (0.2$<$\textit{p}$\leq$0.4), and 15\% are large (\textit{p}$>$0.4). As shown in the first row of Fig.~\ref{fig:example}, the coat in the first clip is displayed in a zoom-in view, where the coat is large and overflows the frame. However, the physical size of the shoes in the third clip is relatively small. 
Because the considerable distance between the shoes and the camera, the scale is visually small.
(2) \textit{Visible duration.}
Due to occlusions and changes in camera perspective, the target product is not always visible in the clip. Here, each clip is categorized by the proportion of visible duration to the entire clip duration, including 48.5\% of long (0.7$<$\textit{p}), 29.6\% of medium (0.4$<$\textit{p}$\leq$0.7) and 21.9\% of short (\textit{p}$\leq$0.4). For example, in the third clip of the fourth row in 
Fig.~\ref{fig:example}, the watch is occluded at the beginning and end of the clip, which significantly increases the difficulty of LPR. Note that the visible duration of the intended product is evaluated by the annotators.
(3) \textit{Background distractor.} 
In the livestreaming of beauty makeup, handbags, and jewelry, \textit{etc.}, there are abundant products displayed on the screen. For example, the first clip in the last row of Fig.~\ref{fig:example} contains more than two dozen perfumes. However, there is only one intended product in a clip, and it is challenging to distinguish the intended product from the distracted background products. Therefore,
we asked the annotators to assess the number (\textit{n}) of products in the clip (or background distractor) and accordingly classify the clips into three subsets, including 13.6\% of abundant (\textit{n}$>$7), 45.8\% of medium (3$<$\textit{n}$\leq$7) and 40.6\% of few (\textit{n}$\leq$3).

\begin{figure*}[htp]
    \begin{center}
    \includegraphics[width=0.85\linewidth]{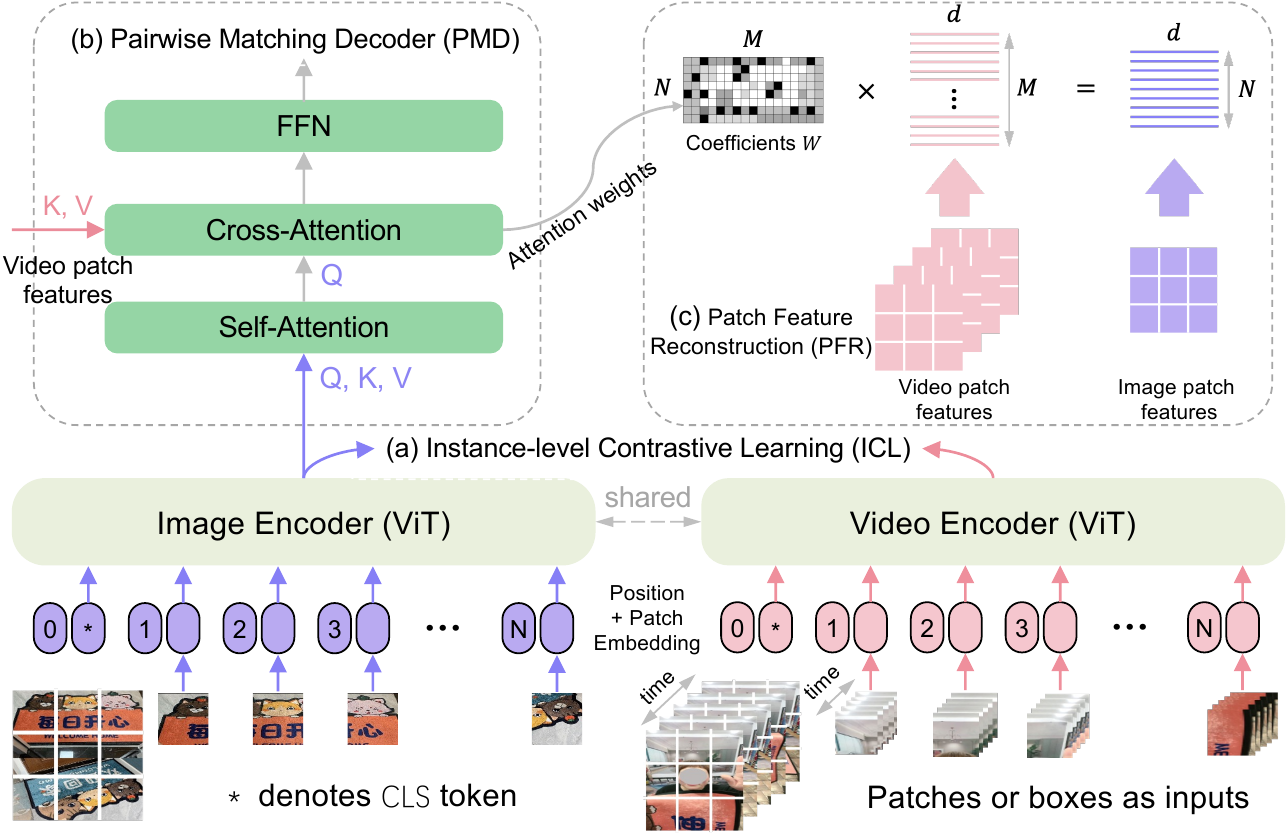}
    \end{center}
    \vspace{-1.0em}
    \caption{Overview of RICE framework. 
    Given two views of the same product, \ie, image and video, 
    the patch features are extracted with the transformer encoder.
    In (a), RICE first performs contrastive loss on the global image and video features. Then, in (b), RICE employs transformer-based
    fusion model to perform patch-level feature interaction between the two views. The optimization of PMD aims to decrease the similarity of the two views from different products and increase that from the same product.
    Furthermore, in (c), RICE exploits reconstruction loss to penalize the misalignment between the semantic patches in the two views, which expects each patch
    feature of the image to be reconstructed from the patch features of the paired video. Best viewed in color.
    }
    \vspace{-1.5em}
    \label{fig:model}
\end{figure*}

\textbf{Clip Description and Image Title.}
In the case of a clip containing multiple products, it is ambiguous for the model
to predict whether the clip and an image match based on visual information only.
Therefore, we additionally provide clip descriptions and image titles to enrich 
the dataset. 
On the one hand, benefiting from promising results achieved by
the Transformer and Convolution Neural Network
based models in ASR~\cite{gulati2020conformer, zhang2020transformer, yeh2019transformer},  
we adopt Conformer~\cite{gulati2020conformer}, a SOTA method on the
widely used LibriSpeech benchmark~\cite{panayotov2015librispeech}, to extract text description from the clip voices. 
On the other hand, the titles of product images are provided by the merchant and are available on the video website.


\subsection{Livestreaming Product Recognition}\label{sec:lpr}
As shown in Fig.~\ref{fig:intro}, this task is to retrieve the GT images from the shop (gallery) for each livestreaming clip (query) and has been considered by several previous works~\cite{huang2015cross, hadi2015buy,liu2016deepfashion,ge2019deepfashion2,cheng2017video2shop, godi2022movingfashion}.
It emphasizes the retrieval performance and considers the
impact of intended product identification. Specifically,
a query clip is counted as missed if the intended product fails to be identified.
Rank-$k$ retrieval accuracy is used to measure
retrieval performance, such that a successful retrieval is counted if the GT image has been retrieved
in the rank-$k$ results.

\textbf{Splitting training and test set.}
In order to evaluate different methods, we split the training and test set and ensure the products in the training and test set are non-overlapping.
The training and test sets contain 4,013,617/332,438 and 20,079/66,358 clips/images, respectively.

\textbf{Intended product box annotation.}
To enable effective supervision of the detector training and evaluation of the detection accuracy, we annotate the intended product box for both the training and test set.
For the training set, we sample 2\% of the clips for intended box annotation and extract 10 frames at even intervals.
For each test clip, we extract one frame every 3 seconds. 
The detection training/test set contains 1,120,410
/501,656 frames with 1,115,629/669,374 intended product boxes, respectively.

\section{Method}
This section presents the technical details of RICE. As shown in Fig.~\ref{fig:model}, the RICE first performs instance-level contrastive 
learning to learn discriminative feature for the product (Sec.~\ref{sec:icl}).
Then, we introduce PMD that pursues fine-grained similarity measurement by 
conducting patch-level feature propagation (Sec.~\ref{sec:pmd}).
The PMD is further guided by the novel PFR loss to promote patch-level semantic alignment (Sec.~\ref{sec:pfr}).
Finally, we study the impact of product location by replacing the input patches with the product boxes produced by intended product detector (Sec.~\ref{sec:ipd}).

\subsection{Instance-level Contrastive Learning (ICL)}\label{sec:icl}
Let $\mathcal{V}$ be a set of livestreaming clips (or videos). Let $\mathcal{I}$
be a set of shop images. The objective of RICE is to learn a function to measure
the similarity between the clip $\mathcal{V}_i\in \mathcal{V}$ and the image $\mathcal{I}_i\in \mathcal{I}$.
Formally, taking $\mathcal{V}_i$ and $\mathcal{I}_i$ as input, the image encoder first splits the image into non-overlapping patches, which are projected into 1D tokens via a linear projection. 
Then the transformer layers are used to extract the patch features, denoted as $\{i_{cls}, i_1, ..., i_N\}$. Likewise, the video encoder processes each video
frame $\mathcal{V}^j_i$ independently and outputs a sequence of video patch
features $\{v_{cls}, v_1, ..., v_M\}$, where $j$ is the index of frame number $|\mathcal{V}_i|$ and $M=N\times|\mathcal{V}_i|$.
Given an image with a resolution of $224\times224$
and patch size of $32\times32$, we have $N=49$.
Note that the image and video encoder share parameters.
Following ViT and CLIP, we extract the global representation from the $\mathsf{[CLS]}$ token.
In order to pull the clip and image of the same product while pushing away that of 
different products in the feature space, we perform InfoNCE loss~\cite{oord2018representation} on the 
global representation, defined as:
\begin{align}\label{loss:nce}
    \mathcal{L}_{nce} = - \mathbb{E}_{p(\Image,\Video)} \left[\log\frac{\exp(g_{\theta}(\Image_i, \Video_i))}{\sum_{\widetilde{\Video}_k\in \widetilde{\Video}} \exp(g_{\theta}(\Image_i, \widetilde{\Video}_k))} \right],
\end{align}
where $g_{\theta}(\Image_i, \Video_i)=g_{\Image}(i_{cls})^\top g_{\Video}(v_{cls})/\tau$ and $\widetilde{\Video}$ consists of a positive sample $\Video_i$ and $|\widetilde{\Video}|-1$ negative samples.
$g_{\Image}$ and $g_{\Video}$ are transformations that map the 
$\mathsf{[CLS]}$ embedding of image and clip, \ie, $i_{cls}$ and $v_{cls}$, to the normalized lower-dimensional features.
$\tau$ is a temperature parameter, and we
use $\tau=0.01$.
The final contrastive loss between the image and clip is a symmetric version of $\mathcal{L}_{nce}$, given by:
\begin{align}\label{loss:cl}
    \mathcal{L}_{c} = - \frac{1}{2}\mathbb{E}_{p(\Image,\Video)} 
    [
    \log\frac{\exp(g_{\theta}(\Image_i, \Video_i))}{\sum_{\widetilde{\Video}_k\in \widetilde{\Video}} \exp(g_{\theta}(\Image_i, \widetilde{\Video}_k))} + \nonumber
    \\
    \log\frac{\exp(g_{\theta}(\Image_i, \Video_i))}{\sum_{\widetilde{\Image}_k\in \widetilde{\Image}} \exp(g_{\theta}(\widetilde{\Image}_k, \Video_i))} 
    ],
\end{align}
where $|\widetilde{\Image}|=|\widetilde{\Video}|$ is the batch size.

 \subsection{Patch-level Semantic Alignment}
\textbf{Pairwise Matching Decoder (PMD).}\label{sec:pmd}
It is straightforward to exploit $g_{\theta}(\Image_i, \Video_i)$ in Eq.~(\ref{loss:nce}) as a measure of the similarity. 
However, the features of the clip and image are extracted independently
from the visual encoder, without information propagation between $\Image_i$ and $\Video_i$. 
To this end, we perform patch-wise feature attention via a transformer decoder layer, named pairwise matching decoder, which consists of a self-attention and a cross-attention layer in this paper. As illustrated in Fig.~\ref{fig:model}~(b), the self-attention layer 
takes the image patch features as the \textit{Query}, \textit{Key} and \textit{Value}, and the cross-attention layer takes the image patch features as \textit{Query} while takes the video
patch features as \textit{Key} and \textit{Value}.
The matching loss of the similarity calculator is defined as follows:
\begin{align}\label{loss:pair}
    \mathcal{L}_{m} = - \frac{1}{2}\mathbb{E}_{p(\Image,\Video)} 
    [
    \log\frac{\exp(f_{\theta}(\Image_i, \Video_i))}{\sum_{\hat{\Video}_k\in \hat{\Video}} \exp(f_{\theta}(\Image_i, \hat{\Video}_k))} + \nonumber
    \\
    \log\frac{\exp(f_{\theta}(\Image_i, \Video_i))}{\sum_{\hat{\Image}_k\in \hat{\Image}} \exp(f_{\theta}(\hat{\Image}_k, \Video_i))} 
    ],
\end{align}
where $f_{\theta}(\Image_i,\Video_i)=v^\top x_{cls}(\Image_i,\Video_i)$, $x_{cls}(\Image_i,\Video_i)$ is the $\mathsf{[CLS]}$ embedding of the decoder layer and $v$ is a
parametric vector. Here, we only sample $N_{neg}$ negative instances for each 
GT $(\Image_i, \Video_i)$ pair, \ie, 
$\hat{\Video}$ consists of a positive sample $\Video_i$ and $N_{neg}$ negative samples.
\begin{figure}[!tp]
    \begin{center}
    \includegraphics[width=0.85\linewidth]{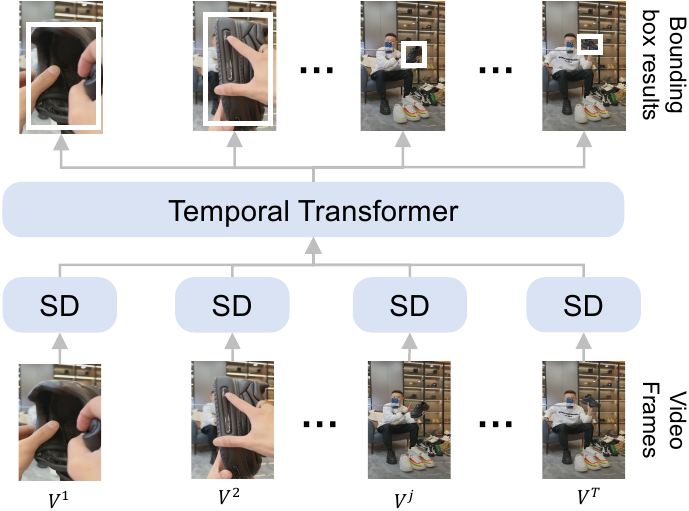}
    \end{center}
    \vspace{-1.0em}
    \caption{Illustration of the Single-Frame Detector (SD) and Multi-Frame Detector (MD). The MD exploits a temporal transformer to fuse 
    the results from the SD.
    }
    \vspace{-1.0em}
    \label{fig:ipd}
\end{figure}

\textbf{Patch Feature Reconstruction (PFR).}\label{sec:pfr}
Furthermore, we pursue cross-view semantic alignment by searching similar patches in the clip to reconstruct the coupled image in the feature space.
Here, we introduce how to perform patch feature reconstruction
given a positive data point of two views of clip $\Video_i$ and image $\Image_i$.
As shown in Fig.~\ref{fig:model}~(b), let 
\begin{align}
X = \{v_1, ..., v_M\} \in \mathbb{R}^{d\times M} \nonumber
\end{align}
be the patch features of the clip, where $v_m\in \mathbb{R}^{d\times 1}$. Likewise, 
the patch features of the image are denoted as:
\begin{align}
Y = \{i_1, ..., i_N\} \in \mathbb{R}^{d\times N}. \nonumber
\end{align}
Then, the $i_n$ can be represented by a linear combination of $X$.
The insight behind this is that the image can be reconstructed from the clip if the clip contains the product in the image.
Therefore, we solve for the coefficients $w_n\in \mathbb{R}^{M\times 1}$ of $i_n$ with respect to $X$.
Finally, the reconstruction
loss is defined as:
\begin{align}\label{loss:recons}
    \mathcal{L}_{r} = || Y-XW ||^2_F.
\end{align}
Since the attention weights $a$ in the cross-attention layer indicate
the correspondences between patches of the two views, 
it is intuitive to learn the reconstruction coefficients $W$ from $a$.
Specifically,
the $a\in \mathbb{R}^{8\times N\times M}$ is fed into two consecutive sets of convolution and
ReLU layers to output the coefficients $W\in\mathbb{R}^{N\times M}$.

The final objective function for the RICE model is the weighted summation 
of $\mathcal{L}_{c}$, $\mathcal{L}_{m}$ and $\mathcal{L}_{r}$, given by:
\begin{align}
    \mathcal{L} = \mathcal{L}_{c}+\mathcal{L}_{m}+\alpha\mathcal{L}_{r},
\end{align}
where $\alpha$ is the trade-off weight and we use $\alpha=0.1$ in the following experiments.

\begin{table*}[htp]
\begin{center}
\begin{tabular} {l|c|ccc|ccc|ccc}
\hline
\multirow{2}{*}{Methods} & \multirow{2}{*}{overall} & \multicolumn{3}{c|}{scale} & \multicolumn{3}{c|}{visible duration} & \multicolumn{3}{c}{number of product} \\
&  & small  &  medium  & large  &  short  &  medium  & long 
&  abundant  &  medium  & few  \\ 
\hline
    ICL$_{patch}$ & 27.1 & 23.9  & \textbf{34.4} & 30.0  & 23.6  & 28.8 & \textbf{36.3} & 20.2  & 25.0 & \textbf{27.5}  \\
    ICL$_{box}$   & 30.0 & 30.1  & \textbf{35.8} & 31.1  & 26.6  & 30.1 & \textbf{37.1} & 17.2  & 29.2 & \textbf{30.6}  \\
\hline
    RICE$_{patch}$ & 31.2& 28.9  & \textbf{37.0} & 32.7  & 28.1 & 32.9 & \textbf{39.6}  & 21.0 & 34.6 & \textbf{31.5} \\
    RICE$_{box}$   & 33.0 & 32.7  & \textbf{39.0} & 33.8  & 29.6  & 34.8 & \textbf{42.0} & 17.6  & 31.9 & \textbf{33.4} \\
\hline
\end{tabular}
\end{center}
\caption{
    The R1 performances of the ICL and RICE model.
    Results of evaluation on different input types, \ie, patch and detected box, are shown in each row.
    The columns show the results on different test subsets split by video variations, \ie, \textit{product scale}, \textit{visible duration}, \textit{number of products}.
    The best performance for each subset is in bold.
}
\label{tab:video-var}
\vspace{-1.0em}
\end{table*}

\subsection{Intended Product Detection (IPD)}\label{sec:ipd}
In order to highlight the intended products in video and suppress the 
background products, we propose replacing patch inputs with the detected 
intended product boxes for the videos. 
As shown in Fig.~\ref{fig:ipd}, we adopt DAB-DETR~\cite{liu2021dab} and TransVOD Lite~\cite{zhou2022transvod} as the SD and MD, respectively. 
The SD detects products frame-by-frame. Given the $T$ frames and one box label per frame, the SD is trained to predict an intended product box for each frame.
However, detecting products with significant appearance changes using only a single frame can be challenging, as exemplified by the \textit{shoe} with \textit{small scale} in $\mathcal{V}^T$ in Fig.~\ref{fig:ipd}. 
Therefore, MD leverages a temporal transformer to capture product interactions in the temporal context and predict a more accurate box for each frame.
For more details about the IPD, please refer to the supplementary material.
\section{Experiment}


\subsection{Dataset and Evaluation Metrics}
Experiments are performed on the LPR4M testset, which contains 20,079 livestreaming clips as query set and 66,358 shop images as gallery set, as described in Sec.~\ref{sec:lpr}.
We adopt rank-\textit{k} accuracy as the retrieval performance metrics.

\subsection{Implementation Details}
\noindent\textbf{Model.}
The image and video encoder share parameters and are
initialized with ViT-B/32 from  CLIP~\cite{radford2021learning}, where the number of layers is 12 and the patch size is 32. Likewise,
we initialize PMD with the similar parameters from CLIP.
\noindent\textbf{Preprocessing.}
We extract 10 evenly spaced frames from each clip
as the video input.
The images and video frames are resized to
$224\times224$. For data augmentation, we randomly mask video frames with a percentage ranging from 0 to 0.9 and a probability of 0.5. 
\noindent\textbf{Optimization.}
We use Pytorch~\cite{paszke2017automatic} to implement the RICE model.
The Adam~\cite{kingma2014adam} optimizer is used with a batch size of 256.
For the learning rate, we decay it using a cosine schedule~\cite{loshchilov2016sgdr} following CLIP. The initial learning rate is 1e-7 for the image encoder and video encoder and 1e-4 for the newly introduced modules.
All the experiments 
are carried out on 
8 NVIDIA Tesla V100 GPUs, which takes about 90 hours
for 3 epochs.

\begin{table}[hpt]
\begin{center}
	\begin{tabular}{r r | ccc}
			
            \hline
			Dataset & Methods & R1 & R5 &R10 \\ 
			\hline
			\parbox[t]{11.5mm}{\multirow{7}{*}{\rotatebox[origin=c]{0}{LPR4M}}}
			& FashionNet~\cite{liu2016deepfashion} & 13.4  & 33.8 & 50.4  \quad \\
			& AsymNet~\cite{cheng2017video2shop} & 22.0  & 46.7 & 63.8  \quad \\					
			& SEAM~\cite{godi2022movingfashion} & 23.3  & 49.5 & 61.4  \quad \\
			
			& NVAN~\cite{liu2019spatially} & 21.4  & 45.2 & 62.7  \quad \\
			& TimeSFormer~\cite{bertasius2021space} & 28.6 & 56.8 & 69.0  \quad \\
			& SwinB~\cite{liu2022video} & 29.1  & 60.1 & 73.9  \quad \\
			
		  	& RICE (Ours) & 33.0  & 65.5 & 77.3  \quad \\
		  	
            
            \hline
            \parbox[t]{6mm}{\multirow{5}{*}{\rotatebox[origin=c]{0}{MF}}}
            & NVAN~\cite{liu2019spatially} & 38.0  & 62.0 & 70.0  \quad \\
            & MGH~\cite{yan2020learning} & 40.0  & 59.0 & 66.0  \quad \\
            & AsymNet~\cite{cheng2017video2shop} & 42.0  & 73.0 & 86.0  \quad \\	
            & SEAM~\cite{godi2022movingfashion} & 49.0  & 80.0 & 89.0  \quad \\
            & RICE (Ours) & 76.1  & 89.7 & 92.6 \quad \\
            \hline
	\end{tabular}
\end{center}
    \caption{
    The LPR4M and MovingFashion (MF) evaluation.
	}
	\label{tab:sota}
	\vspace{-1.45em}
\end{table}

\subsection{Impact of Video Variations}
In this section, we carry out experiments to study the impact of video variations,
\ie, \textit{product scale}, \textit{visible duration}, \textit{number of products}, as shown in Fig.~\ref{fig:example}. 
We evaluate two input types, \ie, patch and detected box, for each model. The results are reported in Table.~\ref{tab:video-var}.
As we can see, the performance declines when
small scale, short visible duration, and abundant products are presented.
(1) Compared to the patch input, the box input significantly improves the accuracy. For
example, ICL$_{box}$ outperforms ICL$_{patch}$ by 6.2\% on \textit{small} split.
Besides, ICL$_{box}$ significantly reduces the performance gap between \textit{small} and 
\textit{medium} split. It indicates the IPD improves the robustness to scale variation.
(2) As the performances on \textit{abundant} split shows, the model with \textit{box} input achieves lower accuracy than \textit{patch} input, because it is challenging for
the detector to distinguish the indent product from the abundant background products. 

\begin{figure}[tp]
    \begin{center}
    \includegraphics[width=0.9\linewidth]{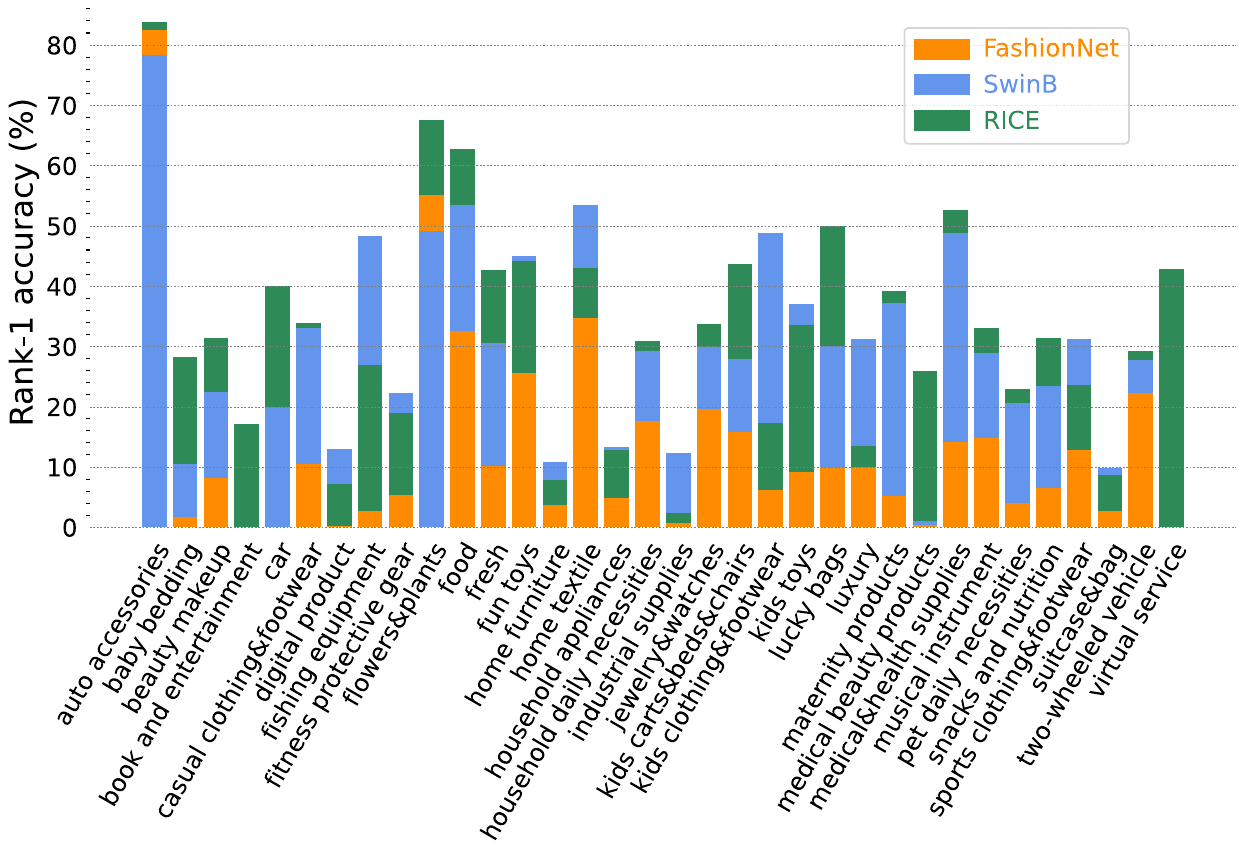}
    \end{center}
    \vspace{-1.2em}
    \caption{Per-category rank-1 performance on the 34 categories.
    }
    \label{fig:cate2acc}
    \vspace{-0.5em}
\end{figure}

\subsection{Comparison with state-of-the-art methods}
In this section, we compare our RICE with state-of-the-art (SOTA) methods 
	on LPR4M and MovingFashion (MF), except AsymNet~\cite{cheng2017video2shop} and WAB because AsymNet is not public available and WAB is a competition dataset with only Chinese introduction. 
	The results are shown in Table~\ref{tab:sota}. 
1) On LPR4M, the FashionNet, AsymNet and SEAM are LPR methods and the others are video understanding methods. As we can see, our RICE surpasses not only the LPR methods but also the strong video understanding methods.
2) On MF, the NVAN and MGH are video understanding methods. Our approach achieves the best accuracy.

\begin{table}[tp]
\begin{center}
	\footnotesize
	\begin{tabular} {c|ccccc|ccc}
            \hline
			\# & \small{ICL} & \small{PMD} & \small{PFR} & \small{IPD} & \small{Txt} &  R1  &  R5  & R10  \\ 
			\hline
            
			a &\checkmark & & & & & 27.1  & 56.4 & 68.3  \quad \\
			b &\checkmark & & & & \checkmark & 28.5  & 58.9 & 71.5  \quad \\
		    c &\checkmark &\checkmark & & & & 29.4  & 62.0 & 73.7  \quad \\
			d &\checkmark &\checkmark &\checkmark & & & 30.3 & 62.7 & 74.0  \quad \\
			e &\checkmark &\checkmark &\checkmark &\checkmark & & 31.3 & 63.2 & 74.3  \quad \\
            f &\checkmark &\checkmark &\checkmark &\checkmark &\checkmark & 33.0 & 65.5 & 77.3  \quad \\
   \hline
		\end{tabular}
	\end{center}
	\vspace{-0.7em}
 \caption{Ablation study on the key components, \ie, ICL: instance-level contrastive learning, PMD: pairwise matching decoder, PFR: patch feature
 reconstruction, IPD: intended product detection, Txt: text modality.
 The rank-\textit{k} accuracy is reported.
	}
	\vspace{-1.0em}
	\label{tab:ablation}
\end{table}

\subsection{Ablation Study}
In this section, we investigate the impact of each component of our approach by conducting ablation experiments.
The results are reported in Table.~\ref{tab:ablation}. 
(c) Compared to the baseline ICL, the PMD obtains the R1 performance gains of 2.3\% (29.4 to 27.1), which demonstrates the superiority of patch-level (local) over instance-level (global) similarity measurement.
(d) The patch-level supervision provided by PFR facilitates semantic alignment and results in a considerable improvement of 0.9\% for R1.
(e) Our IPD replacing patch inputs with detected intended boxes significantly outperforms ICL by 1.0\% R1 as it enables the model to focus on informative regions while suppressing distractions.
In (b) and (f), the addition of text modality increases the R1 from 27.1\% to 28.5\% and 31.3\% to 33.0\%, respectively. It is because the text helps suppress the distracted background products. Here, the ChineseCLIP~\cite{yang2022chinese} is used to extract the embeddings of video ASR and image titles. The text similarity is computed as the dot product of normalized features. Then we combine the text and visual similarities to obtain the final $\langle \mathit{clip}, \mathit{image} \rangle$ similarity via addition.

\subsection{Per-category performance} 

As shown in Fig.~\ref{fig:cate2acc}, we compare the rank-1 accuracy of
FashionNet~\cite{liu2016deepfashion}, SwinB~\cite{liu2022video} and our RICE on all 34 categories. Our RICE consistently outperforms FashionNet on all categories, and outperforms SwinB on most of the categories. 
Due to RICE of averaging the features of frames as 
video features, temporal information is not effectively utilized. 
But SwinB introduces 3D shifted windows to preserve temporal dynamics. 
As a result, SwinB performs well on certain categories with occlusions or view changes, \eg, Suitcase\&bag, as shown in the 5-th row of Fig.~\ref{fig:example}.
The SwinB provides a promising way to enhance our model.


\begin{figure}[tp]
    \begin{center}
    \includegraphics[width=0.9\linewidth]{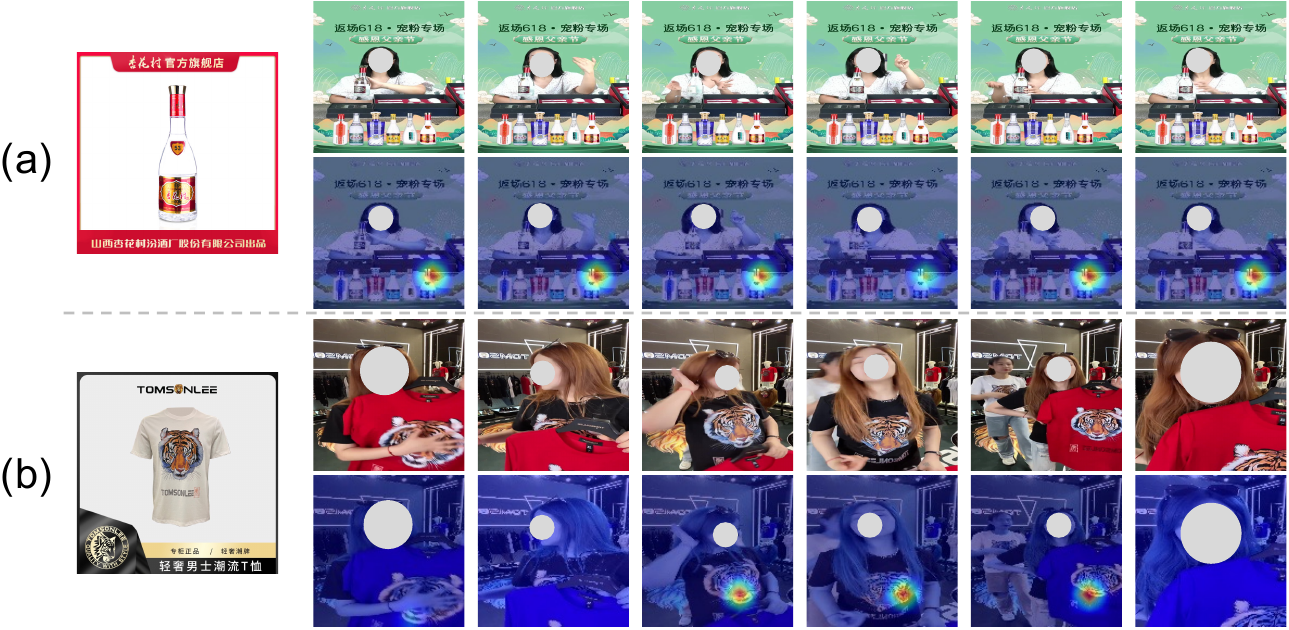}
    \end{center}
    \caption{Attention map visualization of RICE on the LPR4M testset. 
    The first column displays the shop
    images. We show the raw frames and the corresponding attention maps for each video.
    }
    \vspace{-1.2em}
    \label{fig:goodcase}
\end{figure}

\subsection{Attention Region Visualization}
To provide insight into PMD, we conduct further visualization. 
In Fig.~\ref{fig:goodcase}, we show the attention map of RICE$_{patch}$
between shop image and video patches, where an image is regarded as the query, and attention weights
on all spatial patches are visualized. We use the attention weights in the 
cross-attention layer of PMD for visualization. We make the following observations.
(1) For the complex scenarios like (a) in Fig.~\ref{fig:goodcase}, 
our approach can distinguish the target \textit{Chinese liquor} from the nearby background \textit{liquors}.
(2) Interestingly, as shown in (b) of Fig.~\ref{fig:goodcase}, even the target product is not always visible in the video,
our approach still focuses on the corresponding regions accurately while pays 
less attention to the occluded regions.

\section{Conclusions}
In this paper, we present a large-scale dataset that offers broader coverage of categories and more sufficient data modalities named LPR4M. 
Moreover, the RICE model is proposed to integrate instance-level contrastive 
learning and patch-level cross-view semantic alignment mechanism into a framework. 
The extensive experiments demonstrate the effectiveness of the proposals clearly and show that additional performance gains can be achieved via integrating 
intended product detection and text modality. 
In this work, we show that it is a promising way to enhance the LPR model 
from the aspect of large-scale multimodal training.
We hope the proposed LPR4M and the RICE baseline can spur further investigation into the LPR task.

{\small
\bibliographystyle{ieee_fullname}
\bibliography{egbib}
}

\end{document}